\pdfoutput=1

\documentclass[11pt,table,xcdraw]{article}

\usepackage[]{ACL2023}

\usepackage{times}
\usepackage{latexsym}

\usepackage[T1]{fontenc}

\usepackage[utf8]{inputenc}

\usepackage{microtype}

\usepackage{inconsolata}

\usepackage{booktabs}
\usepackage{multirow}
\usepackage{amsmath}
\usepackage{graphicx}
\usepackage{color}
\usepackage{xcolor}
\usepackage{algorithm}
\usepackage{algorithmicx}
\usepackage{algpseudocode}
\usepackage{pifont}
\usepackage{booktabs}
\usepackage{CJKutf8}

\begin{document}
\begin{CJK}{UTF8}{gbsn}

\renewcommand{\thefootnote}{\alph{footnote}}

\title{BLEURT Has Universal Translations: An Analysis of Automatic Metrics \\ by Minimum Risk Training}

\author{
    Yiming Yan\textsuperscript{\rm 1}\protect\footnotemark[1]~, 
    Tao Wang\textsuperscript{\rm 2}, 
    Chengqi Zhao\textsuperscript{\rm 2}, 
    Shujian Huang\textsuperscript{\rm 1}\protect\footnotemark[2]~, \\
    \textbf{Jiajun Chen\textsuperscript{\rm 1}}, 
    \textbf{Mingxuan Wang\textsuperscript{\rm 2}} \\
    \textsuperscript{\rm 1} National Key Laboratory for Novel Software Technology, Nanjing University, China \\
    \textsuperscript{\rm 2} ByteDance AI Lab, China\\
    \texttt{yanym@smail.nju.edu.cn}, ~\texttt{\{huangsj, chenjj\}@nju.edu.cn} \\
    \texttt{\{wangtao.960826, zhaochengqi.d, wangmingxuan.89\}@bytedance.com}
}

\maketitle

\renewcommand{\thefootnote}{\fnsymbol{footnote}}
\footnotetext[1]{Work was done during internship at ByteDance AI Lab.}
\footnotetext[2]{Corresponding author.}

\renewcommand{\thefootnote}{\arabic{footnote}}

\begin{abstract}
Automatic metrics play a crucial role in machine translation. Despite the widespread use of n-gram-based metrics, there has been a recent surge in the development of pre-trained model-based metrics that focus on measuring sentence semantics. However, these neural metrics, while achieving higher correlations with human evaluations, are often considered to be black boxes with potential biases that are difficult to detect. In this study, we systematically analyze and compare various mainstream and cutting-edge automatic metrics from the perspective of their guidance for training machine translation systems. Through Minimum Risk Training (MRT), we find that certain metrics exhibit robustness defects, such as the presence of universal adversarial translations in BLEURT and BARTScore. In-depth analysis suggests two main causes of these robustness deficits: distribution biases in the training datasets, and the tendency of the metric paradigm. By incorporating token-level constraints, we enhance the robustness of evaluation metrics, which in turn leads to an improvement in the performance of machine translation systems. Codes are available at \url{https://github.com/powerpuffpomelo/fairseq_mrt}.

\end{abstract}

\section{Introduction}
Automatic metrics are crucial for the training of machine translation models, as they can measure translation quality at low cost.
Currently, the most widely used translation evaluation metric is still the n-gram-based BLEU \citep{papineni2002bleu, marie2021scientific}. However, it is acknowledged that BLEU, which relies on the surface-level vocabulary matching, exhibits significant limitations \citep{smith2016climbing, reiter2018structured, mathur2020tangled, kocmi2021ship}. For instance, BLEU fails to differentiate between errors of varying severity and assigns equal weight to each word.

\begin{figure}[h]
  \centering
  \includegraphics[width=0.48\textwidth]{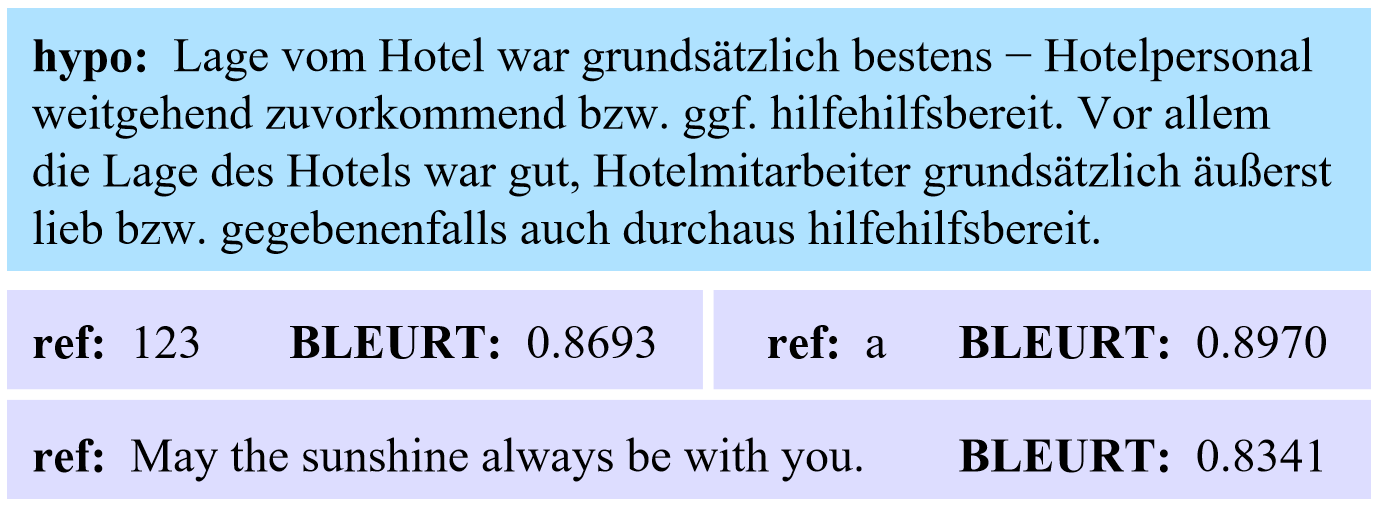}
  \caption{An example of a universal adversarial translation of BLEURT. $hypo$ means the translation sentence and $ref$ means the reference sentence. BLEURT needs to compare $hypo$ and $ref$ to judge the quality of $hypo$. This figure shows that the universal translation can achieve high BLEURT scores when calculated with each $ref$, even if $hypo$ and $ref$ are completely unrelated.}
  \label{fig:bleurt_universal_translation_example}
\end{figure}

In recent years, the advent of pre-trained models \citep{devlin2018bert, liu2019roberta, conneau2019unsupervised, yang2019xlnet, lan2019albert} has led to significant advancements in the development of metrics such as BLEURT \citep{sellam2020bleurt} and COMET \citep{rei2020comet}, which employ pretrained language models (PLM) to assess the semantic meaning of sentences. These approaches have been shown to outperform metrics that rely on superficial word matching and have a more consistent correlation with human annotation. Despite these advances, it is important to note that neural metrics are characterized by opaque decision bases and may be subject to biases that are more difficult to detect \citep{sun2022bertscore}.  
Therefore, we aim to conduct an analysis of the properties of various metrics  in order to gain a deeper understanding.
While there have been recent studies on the analysis of metrics \citep{kocmi2021ship, hanna2021fine, sun2022bertscore}, these works primarily focus on examining metric scores on specific datasets. To the best of our knowledge, this paper is the first to analyze metrics from the perspective of their guidance for training machine translation systems.

\begin{table*}[]
\centering
\begin{tabular}{ccccc}
\toprule[1.2pt]
\textbf{Metrics} & \textbf{Supervised} & \textbf{Paradigm} & \textbf{Based PLM}                   & \textbf{Considered input forms}                                          \\ \hline
BLEU             & \ding{55}                   & Match             & -                                    & \textless{}hyp, ref\textgreater{}                                        \\
BERTScore        & \ding{55}                   & Match             & RoBERTa ~/ BERT  & \textless{}hyp, ref\textgreater{}                                        \\
BARTScore        & \ding{55}                   & Generation        & BART                                 & \textless{}hyp, ref\textgreater{} ~/ \textless{}src, hyp\textgreater{}                                       \\
BLEURT           & \ding{51}                   & Regression        & BERT                                 & \textless{}hyp, ref\textgreater{}                                        \\
COMET            & \ding{51}                   & Regression        & XLM-RoBERTa                          & \textless{}hyp, src, ref\textgreater{}                                   \\
UniTE            & \ding{51}                   & Regression        & XLM-RoBERTa                          & \textless{}hyp, ref\textgreater ~/ \textless{}hyp, src, ref\textgreater{} \\ \bottomrule[1.2pt]
\end{tabular}
\caption{Summary of metrics considered in this paper.}
\label{tab:metrics_info}
\end{table*}

In this paper, we employ Minimum Risk Training (MRT) \citep{shen2015minimum} to train translation models.
Compared to Maximum Likelihood Estimation (MLE), MRT can reduce the gap between training and evaluation, resulting in higher quality translations \citep{shen2015minimum, edunov2017classical}. In addition, since MRT uses metrics to optimize translation models, we can explore the impact of metrics on translation by observing the MRT training process.

Our experiment results show that MRT reveals the robustness defects in some metrics:
the training collapses and the generated translations, despite getting high metric scores, show poor translation quality. 
For instance, we find universal adversarial translations of BLEURT and BARTScore, which are capable of obtaining high scores when evaluated against any reference sentence.
An example is presented in Figure~\ref{fig:bleurt_universal_translation_example}.
Further analysis shows that the robustness defects are rooted in the distribution biases of the training corpora, as well as in the tendency of the metric modeling paradigm.
In addition, we explore methods for optimizing metrics and translation models: word-level information constraints are introduced by combining MRT with NLL loss and metric ensemble. 



Our main contributions are as follows:
\begin{itemize}
\item We present a systematic analysis of automatic metrics for machine translation from the perspective of guidance for training machine translation systems.

\item We provide analytical conclusions, including metric robustness deficiencies, as well as an analysis of the underlying causes.

\item We explore methods to improve metric robustness and translation quality and demonstrate their effectiveness. 

\end{itemize}

\section{Analyze Metrics with MRT}
We train translation models in two stages: in the MLE training phase, the model is trained with conventional negative log-likelihood (NLL) loss; then in the MRT training phase, we fine-tune the model with each metric, so as to obtain translation models with various metric styles. In this way, the characteristics of different metrics can be analyzed through observing the changes in the training process and the translation results.

\subsection{Considered metrics}
Given the translated sentence $hyp$, the automatic evaluation metric evaluates $hyp$ by comparing it with the reference sentence $ref$ (and sometimes with the source sentence $src$).
This paper selects the most mainstream and cutting-edge six metrics for comparison and analysis, including three unsupervised metrics: BLEU \citep{papineni2002bleu}, BERTScore \citep{zhang2019bertscore}, BARTScore \citep{yuan2021bartscore}, and three supervised metrics: BLEURT \citep{sellam2020bleurt}, COMET \citep{rei2020comet}, UniTE \citep{wan2022unite}. The specific information is shown in Table~\ref{tab:metrics_info}. 

We use SacreBLEU\footnote{\url{https://github.com/mjpost/SacreBLEU}} and F1-score\footnote{\url{https://github.com/Tiiiger/bert_score}} as a measure of text quality to calculate BLEU and BERTScore respectively. Following the instructions of \citet{yuan2021bartscore}, we use the CNNDM version of BARTScore\footnote{\url{https://github.com/neulab/BARTScore}} to calculate the F1-score of $\langle hyp, ref \rangle$ for translate-to-English language pairs, and multilingual BART to obtain the faithfulness by calculating $P(hyp \mid src)$ for the other language pairs.
As recommended, we use BLEURT-20\footnote{\url{https://github.com/google-research/bleurt}} and WMT20-COMET-MQM\footnote{\url{https://github.com/Unbabel/COMET}} to compute BLEURT and COMET respectively. For UniTE, since our task is multilingual, we use UniTE-MUP\footnote{\url{https://github.com/NLP2CT/UniTE}} in our experiments.
It is worth noting that, for a fair comparison, we consider two input forms of UniTE: one uses $\langle hyp, ref \rangle$ to calculate the translation quality, which we denote as UniTE\_ref; the other uses $\langle src, hyp, ref \rangle$, which we denote as UniTE\_src\_ref.

\subsection{Minimum Risk Training}
Minimum Risk Training (MRT) is a sequence-level objective that aims to minimize the expected risk on the training data.
Given a training set $\mathcal{D}=\{(\text{x,y})\}$, MRT uses the loss function $\Delta(\hat{\text{y}},\text{y})$ to compute the discrepancy between the ground truth $\text{y}$ and the model prediction $\hat{\text{y}}$.

Different from conventional MLE training methods, MRT allows the use of arbitrary non-differentiable loss functions. 
Therefore, automatic metrics can be introduced to train machine translation systems.
While an MLE-trained model may not translate authentically, MRT can produce more natural translation results by reducing the gap between training and evaluation \citep{shen2015minimum, edunov2017classical, wang2020exposure}.

In MRT training, risk is defined as the expected loss with respect to the posterior distribution:
\begin{align}
\mathcal{R}(\theta) = \sum_{(\text{x,y})\in \mathcal{D}}\sum_{\hat{\text{y}}\in \mathcal{Y}(\text{x})} P(\hat{\text{y}}|\text{x};\theta)\Delta(\hat{\text{y}},\text{y})
\end{align}
in which $\mathcal{Y}(\text{x})$ is the set of all possible translations of $\text{x}$.
Since the full search space is intractable, we choose a certain number of candidate translations as a subset to approximate the posterior distribution.

\begin{table}[]
\centering
\begin{tabular}{cccc}
\toprule[1.2pt]
               & \textbf{Train} & \textbf{Valid} & \textbf{Test} \\ \hline
\textbf{En$\Leftrightarrow$De} & 4.3M           & 3000           & 3003          \\
\textbf{En$\Leftrightarrow$Zh} & 1.3M           & 1797           & 4534          \\
\textbf{En$\Leftrightarrow$Fi} & 2.5M           & 2500           & 2507          \\ 
\bottomrule[1.2pt]
\end{tabular}
\caption{Statistics of datasets on three language pairs.}
\label{tab:dataset}
\end{table}

\subsection{Experiment Setup}

\paragraph{Dataset}
With reference to datasets and language pairs that are widely used in machine translation and neural metrics studies, we conduct experiments on six language directions: English-German (En$\Leftrightarrow$De), English-Chinese (En$\Leftrightarrow$Zh), English-Finnish (En$\Leftrightarrow$Fi).
We use the WMT14 training corpus for En$\Leftrightarrow$De, and the newstest13 and newstest14 are the validation set and the test set, respectively.
For En$\Leftrightarrow$Zh, we use the LDC corpus as training data, and the NIST 2002, 2003 are used for validation, while NIST 2004, 2005, 2006 are used as the test sets.
For En$\Leftrightarrow$Fi, the datasets are from the training-parallel-ep-v8 and rapid2016 sections of WMT17, where the validation set and the test set are split at a rate of 0.1\% respectively.
The statistics of the datasets are shown in Table~\ref{tab:dataset}.

\paragraph{Implentation Details}
We train Transformer Base setting \citep{vaswani2017attention} using the fairseq\footnote{\url{https://github.com/pytorch/fairseq}} toolkit, where the model consists of 6 layers of encoder and 6 layers of decoder with hidden size of 512.
In the MLE training phase, the batch size is 65,536. The best checkpoint is selected based on the BLEU scores on the validation set.
For evaluation, we average the last ten checkpoints and use beam search for inference.
In the MRT training phase, each batch contains 8,000 tokens. Following previous work on MRT \citep{edunov2017classical}, we use beam search to generate candidates, and the beam size is set to 12. The best checkpoint is selected based on the corresponding metric.
We list the training duration for MLE and MRT in Appendix \ref{sec:train_duration}.
For all language pairs, sentences are encoded using byte pair encoding \citep{sennrich2015neural} with 32,000 merge operations, jointly learned from both the source and target side of the training data.
We use Adam~\citep{kingma2014adam} optimization and the same learning rate schedule as described in \citet{vaswani2017attention} with the warm-up step of 4,000.

\begin{table*}[]
\centering
\begin{tabular}{ccccccc}
\toprule[1.2pt]
             & \textbf{En$\Rightarrow$De}            & \textbf{De$\Rightarrow$En}             & \textbf{En$\Rightarrow$Zh}            & \textbf{Zh$\Rightarrow$En}             & \textbf{En$\Rightarrow$Fi}            & \textbf{Fi$\Rightarrow$En}                        \\ \hline
MLE Training & \cellcolor[HTML]{FFFFFF}28.4 & \cellcolor[HTML]{FFFFFF}31.4 & \cellcolor[HTML]{FFFFFF}37.2 & \cellcolor[HTML]{FFFFFF}45.4 & \cellcolor[HTML]{FFFFFF}28.7 & \cellcolor[HTML]{FFFFFF}38.1 \\ 
\bottomrule[1.2pt]
\end{tabular}
\caption{SacreBLEU scores on the test sets obtained by training Transformer-base with MLE.}
\label{tab:mle_results}
\end{table*}

\begin{figure*}[h]
  \centering
  \includegraphics[width=\textwidth]{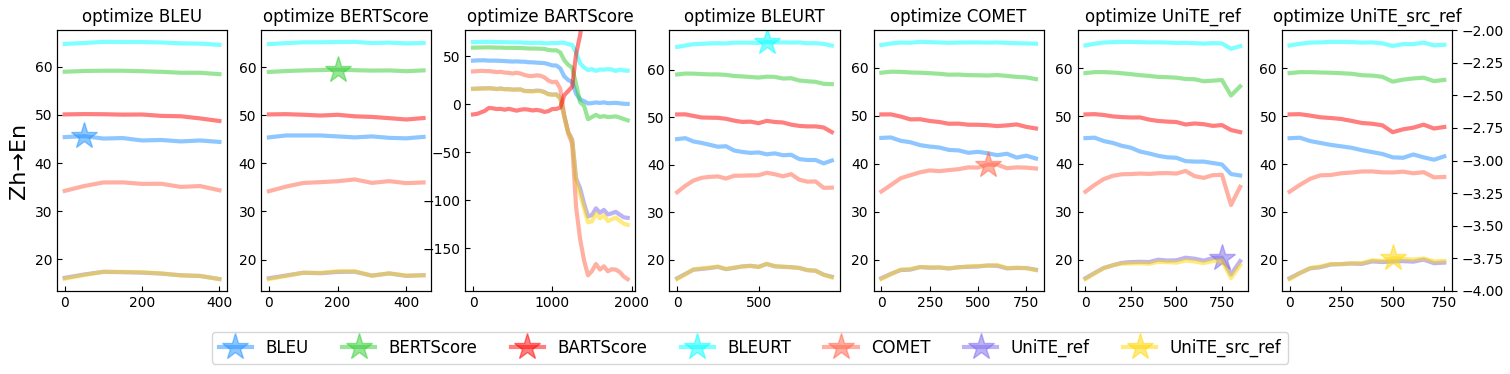}
  \caption{The training process of MRT optimized by each metric on Zh$\Rightarrow$En. The horizontal axis represents the training steps, and the vertical axis is the score of each metric  (except for BARTScore on the right axis, which is a negative number because it calculates the logarithmic probability of translations); metrics other than BARTScore and BLEU are mostly distributed between 0 and 1, and we multiply them uniformly by 100 for ease of observation. The asterisk represents the highest value achieved by the optimized metric.}
  \label{fig:mrt_zh2en}
\end{figure*}

\subsection{Main Results}
The MLE stage is the main factor in improving translation performance of the model, whereas MRT fine-tuning directs the model towards specific metrics.
The SacreBLEU scores of the translation models after MLE training are shown in Table~\ref{tab:mle_results}.
Then in the MRT fine-tuning phase, we use six metrics separately on each language pair to guide the training. 
Figure~\ref{fig:mrt_zh2en} shows the evaluation results of the translations by optimizing each metric on Zh$\Rightarrow$En during this phase \footnote{Due to space limitations, please refer to Appendix \ref{MRT Training Process Figures} for the complete graph of training states on each metric and language pair.}.

We investigate the changes in the MRT curve for each metric and language pair. The remaining of the metrics generally improve along with the optimized metrics, followed by a slight decrease, indicating that there are differences in the quality evaluation criteria of different metrics. In general, all metrics remain basically stable during the MRT process. 

However, we find several exceptions, such as optimizing BLEURT on the En$\Rightarrow$De and En$\Rightarrow$Zh language pairs, where the rest of the metrics experience a severe drop. As shown in Table~\ref{tab:mrt_stat_en2de}, BLEURT remains basically stable, but the rest of the metrics drop to particularly low or even negative values. The same situation occurs when optimizing BARTScore, as shown in  Table~\ref{tab:mrt_stat_en2de} and Figure~\ref{fig:mrt_zh2en}.

\paragraph{MRT Exposes the Robustness Defects of Metrics}
We find deficiencies in some metrics when MRT collapses. For example, we find that there are universal adversarial translations in both BLEURT and BARTScore.

\textbf{(1) Universal translations of BLEURT.}
We take the checkpoint of the translation model on  En$\Rightarrow$De where BLEURT reaches the highest point to generate translations on the test set. The decoded results show that the translation quality does indeed collapse severely.
Table \ref{tab:decode_trans_stat} shows the two most frequently decoded translations.
It can be seen that the translation model generates many similar sentences with high frequency, regardless of the source sentences. This shows that decoding such sentences can get high BLEURT scores. The example of calculating the BLEURT scores of universal translations is also shown in Figure \ref{fig:bleurt_universal_translation_example}.

\textbf{(2) Universal translations of BARTScore.}
We also generate translations with the checkpoint on De$\Rightarrow$En which gets highest BARTScore. As shown in Table \ref{tab:decode_trans_stat}, the translation model also decodes many similar sentences, but unlike BLEURT, the form of the high-frequency decoded sentences is only repetition of simple words.


The phenomenon of universal adversarial translations shows that BLEURT and BARTScore are flawed, and a high metric score does not mean high translation quality.
If the metric is not good enough, it actually leads the translation model in the wrong direction.

\begin{table*}[]
\small
\centering
\begin{tabular}{cccccccc}
\toprule[1.2pt]
\multirow{2}{*}{\textbf{Optimized Metric}} & \multicolumn{7}{c}{\textbf{Change Range of Metrics During MRT on En$\Rightarrow$De}} \\
 & BLEU & BERTScore & BARTScore & BLEURT & COMET & UniTE\_ref & UniTE\_src\_ref \\ \hline
BLEU & 0.00\% & 0.00\% & 0.00\% & 0.00\% & 0.00\% & 0.00\% & 0.00\% \\
BERTScore & 0.70\% & 0.69\% & -0.24\% & 0.71\% & 4.61\% & 4.31\% & 3.98\% \\
BARTScore & -100.00\% & -176.84\% & 92.15\% & -79.78\% & -574.39\% & -397.07\% & -385.80\% \\
BLEURT & -100.00\% & -107.48\% & -20.33\% & 14.96\% & -435.00\% & -423.12\% & -408.72\% \\
COMET & -14.79\% & -3.01\% & -0.92\% & 1.65\% & 13.38\% & 10.86\% & 9.90\% \\
UniTE\_ref & -31.69\% & -11.51\% & -3.12\% & -0.37\% & 2.66\% & 19.11\% & 18.06\% \\
UniTE\_src\_ref & -39.08\% & -15.27\% & -4.16\% & -2.39\% & -4.28\% & 21.99\% & 21.72\% \\ \bottomrule[1.2pt]
\end{tabular}
\caption{The change range of all metrics when one metric is optimized to the highest value during MRT on En$\Rightarrow$De.
$0.00\%$ means that the optimized metric does not continue to improve, and the highest value remains the same as the result of MLE training; a negative number means that the metric score goes from positive to negative, which means it decreases a lot. (For the results of the remaining five language directions, see Appendix \ref{MRT Training Process Statistics})}
\label{tab:mrt_stat_en2de}
\end{table*}

\begin{table*}[]
\centering
\small
\begin{tabular}{ccc}
\toprule[1.2pt]
                                                                                                         & \textbf{Frequency} & \textbf{Decoded Translations with Top2 Frequency}                                                                                                                                                                                                                                                    \\ \hline
\multirow{2}{*}{\begin{tabular}[c]{@{}c@{}}Optimize \\ BLEURT on \\ En$\Rightarrow$De\end{tabular}}    & 689                & \begin{tabular}[c]{@{}c@{}}Lage vom Hotel war grundsätzlich bestens − Hotelpersonal weitgehend  zuvorkommend  \\ bzw.  ggf. hilfehilfsbereit. Vor allem die Lage des Hotels  war gut, Hotelmitarbeiter \\ grundsätzlich  äußerst lieb bzw. gegebenenfalls auch durchaus hilfehilfsbereit.\end{tabular} \\ \cline{2-3} 
                                                                                                         & 386                & \begin{tabular}[c]{@{}c@{}}Lage vom Hotel war grundsätzlich bestens − HotelPersonal weitgehend  zuvorkommend \\ bzw.   ggf. hilfehilfsbereit. Vor allem die Lage des Hotels  war gut, Hotelmitarbeiter \\ grundsätzlich  äußerst  lieb bzw. gegebenenfalls auch durchaus hilfehilfsbereit.\end{tabular} \\ \hline
\multirow{2}{*}{\begin{tabular}[c]{@{}c@{}}Optimize \\ BARTScore on \\ De$\Rightarrow$En\end{tabular}} & 141                & \begin{tabular}[c]{@{}c@{}}! Mallorca! Mallorca! Mallorca! Mallorca! Mallorca! Mallorca! Mallorca!  Mallorca! \\ Mallorca!  Mallorca! Mallorca! Mallorca! Mallorca! Mallorca!  Mallorca! Mallorca! \\ Mallorca! Mallorca!\end{tabular}                                                                  \\ \cline{2-3} 
                                                                                                         & 137                & \begin{tabular}[c]{@{}c@{}}Mallorca! Mallorca! Mallorca! Mallorca! Mallorca! Mallorca! Mallorca!  Mallorca! \\  Mallorca! Mallorca! Mallorca! Mallorca!\end{tabular}                                                                                                                                   \\ 
\bottomrule[1.2pt]
\end{tabular}
\caption{Examples of decoded translations of BLEURT and BARTScore. Due to space limitations, only the top2 frequency translations are listed.}
\label{tab:decode_trans_stat}
\end{table*}

\subsection{Analysis}
\subsubsection{Why Universal Translations Exist}
We examine the WMT14 En$\Leftrightarrow$De parallel corpora, and find that there are many sentences with similar semantics in   the training set, including a large corpus of hotel reviews that are semantically similar to universal translations of BLEURT \footnote{Some examples can be found in Appendix \ref{high_freq}.}. 
This indicates that the patterns of  universal translations are related to the translation training set, and they  come from the high frequency samples in the training corpora.
\citet{raunak2021curious} also mentions the problem of corpus bias, whose study on NMT hallucinations shows that specific noise patterns in the training corpora lead to specific hallucination patterns.
Due to high frequency samples in the translation training set, it becomes easy for the translation model to decode certain sentences (even if they have nothing to do with the source sentences).

Moreover, the high score of the metric condones the model to decode such sentences, leading to the creation of universal  translations.
BLEURT uses metric data and generates a large amount of pseudo-data for supervised training, and the metric data comes from the translation training corpus. Since data augmentation may introduce noise and amplify hallucinations \citep{raunak2021curious}, we suggest that its indulgence of universal translations is also related to the training corpus.

The universal translations of BARTScore contain repetitions of simple words, which is similar to the hallucination phenomena that occurs in the early stages of translation model training \citep{voita2021language}.
We not only use the $F1$ score, but also experiment with the $Recall$ of BARTScore (computing $P(ref|hyp)$) to guide the training, and find that this setting can produce universal suffixes, that is, even if the correct translation is followed by a specific suffix, it does not reduce the BARTScore. Therefore, we suggest that the vulnerability of BARTScore may be due to the fact that it uses model generation probabilities to determine translation quality, and this generation-based metric tends to assign high scores to easily generated sentences during translation model training.
In short, the defects may stem from the tendency of the metric modeling paradigm \footnote{Metrics can be categorized into different modeling paradigms, including matching, regression, generation, and so on \citep{sun2022bertscore,yuan2021bartscore}.}.

The phenomenon of universal adversarial translations suggests that, on the one hand, we need to optimize the translation and metric datasets to balance their distributions, avoiding high-frequency samples; on the other hand, we need to optimize the metrics so that they are as little affected by the distribution bias of the dataset as possible. For example, sentence-level metrics can be constrained by incorporating word-level information. We present this experiment in Section \ref{sec:Optimize Metrics and Translations}.

\subsubsection{Comparison of Metrics}
We observe and compare the changes in the training effect of translation models guided by each metric on each language pair, and the summary is as follows:

BLEU converges quickly. This is as expected, since the translation model is selected by BLEU in the general MLE training phase, there is almost no continuous optimization during  MRT.
BERTScore also converges in a few steps. 
When BERTScore is optimized, other metrics remain relatively stable and sometimes show an upward trend.

The consistency between BLEURT and other metrics shows language pair differences: for translate-to-English language pairs, the other metrics change steadily and show high consistency with BLEURT. All three to-En language pairs show an increase in COMET, UniTE\_ref, and UniTE\_src\_ref. However, on the language pairs that translate from English, the consistency becomes very poor, where the other metrics drop significantly when optimizing translation models with BLEURT.

The metric that is least consistent with other metrics is BARTScore. On all language pairs, the rest of the metrics decrease when BARTScore is used to train translation models. 

COMET, UniTE\_ref, and UniTE\_src\_ref are similar and  can  improve each other. However, when optimizing with these metrics, a decrease in BLEU is observed for all language pairs. This may indicate that the translation model is gradually trained to be more inclined towards translations that are semantically close to the reference sentences, but the specific words may not be the same.  
In addition, other metrics also show a smooth trend of change, indicating that these metrics may be superior and more robust.

\begin{figure}[h]
  \centering
  \includegraphics[width=0.48\textwidth]{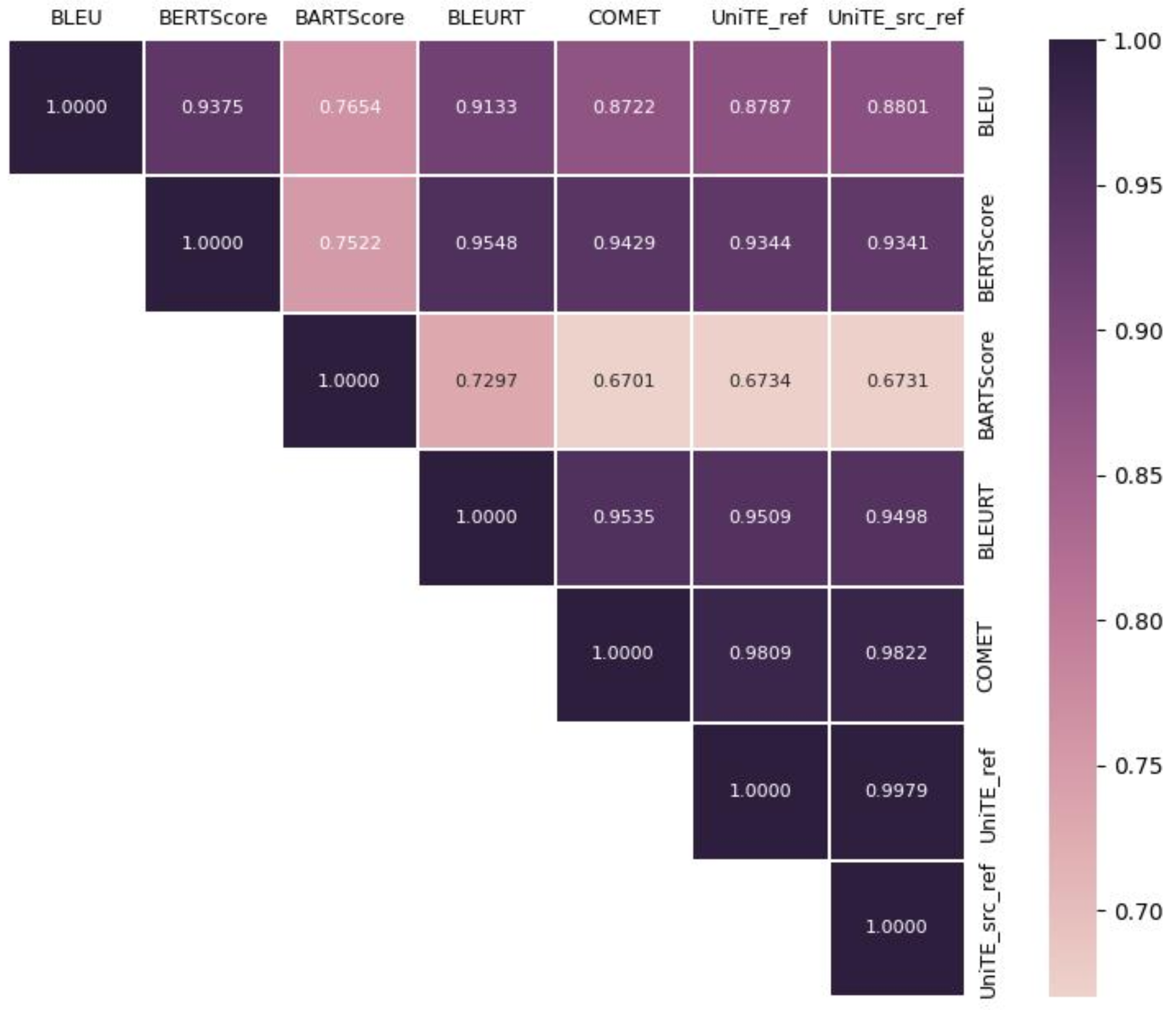}
  \caption{Pairwise correlation of metrics on translate-to-English language pairs. All metrics are significantly positively correlated ($p<0.001$). Ignoring self-relevance, the correlations between COMET and UniTE, BLEURT and BERTScore are particularly strong.}
  \label{fig:metric_correlation}
\end{figure}

\paragraph{Same Pre-trained Model Leads to Similar Metrics.}
We also find a pattern that metrics that are based on the same pre-trained model have similar trends in the variation of the training effect of MRT.  
We count the pairwise correlation of each metric, and find that the correlation between BERTScore and BLEURT (both based on BERT), and the correlation between COMET and UniTE (both based on XLM-Roberta) are higher than other metric pairs for translate-to-English language pairs, as shown in Figure \ref{fig:metric_correlation}.
For language pairs translated from English, the robustness bias of BLEURT weakens its correlation with BERTScore, but the Pearson correlation coefficient still reaches $0.82$ and is significantly correlated.
This indicates that metrics based on the same pre-trained model have more consistent criteria for the evaluation of translation quality. 


\paragraph{Robust Metrics can Drive Improvement in Other Metrics.}
MRT experiments show that the optimization process of BARTScore as well as BLEURT (on translation-from-English language pairs) is accompanied by a strong decrease of the other metrics, and we find metric robustness deficits in these cases. Therefore, we suggest that robust metrics may drive other metrics to improve together during MRT.  (However, the converse inference does not hold. The ability to drive other metrics to improve is not sufficient to conclude that the metrics are robust enough, because metrics may have common deficits that have not yet been discovered.)

\begin{figure*}[h]
  \centering
  \includegraphics[width=\textwidth]{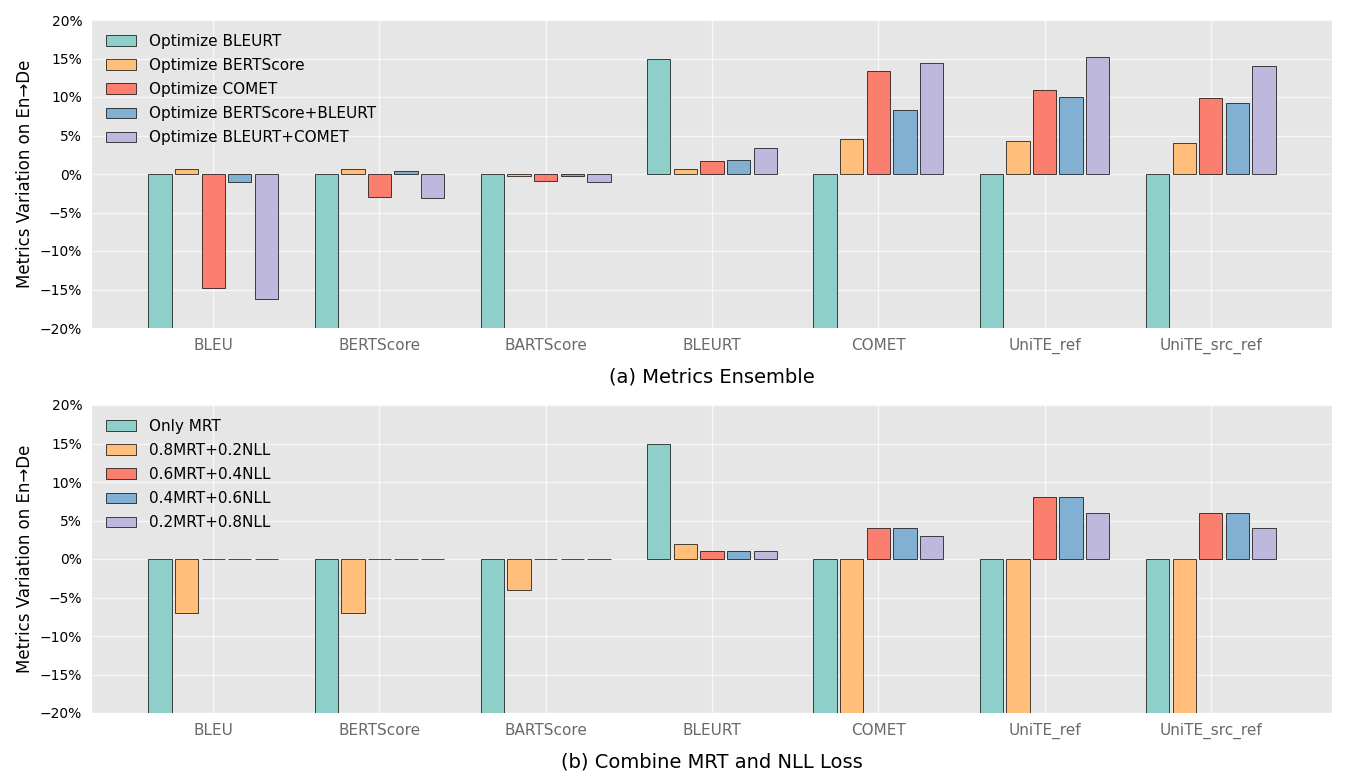}
  \caption{This figure displays two improvement strategies (<a> metrics ensemble and <b> modifying loss), with each color representing a different training approach. The figure is divided into seven groups from left to right, representing the range of change in each metric after training with a particular approach. For instance, the green bar represents the impact of utilizing BLEURT to guide translation model training in MRT, where only BLEURT improves while the other metrics decline significantly (we only display the range of -20\% to 20\% ), which is consistent with the previous charts.}
  \label{fig:optimize}
\end{figure*}

\section{Optimize Metrics and Translations}
\label{sec:Optimize Metrics and Translations}
The analysis of the MRT training process allows us to understand the impact of each metric on translation quality. Our goal is both to exploit the advantages of the MRT training approach and to avoid training collapse due to the robustness deficiencies of the metrics.

MRT needs to sample many translation sentences in advance, and then use sentence-level metrics to predict the scores and calculate the loss. If the metrics that guide translation training do not take word-level information into account, the translation model may ignore details and gradually deviate during the training process.  
Therefore, we try two methods to constrain the training direction by introducing word-level constraints: combining MRT and NLL loss, and doing metrics ensemble.




\subsection{Combine MRT and NLL Loss}
\subsubsection{Experiments}
We take the fine-grained word-level similarity as a part of  the objective function by incorporating the NLL loss, which computes the $log$ loss for each token.
We set the hyperparameter $\lambda_{MRT}$ to control the weights. The formula is as follows:
\begin{align}
\mathcal{L} = {\lambda}_{MRT} * \mathcal{L}_{MRT} + (1-{\lambda}_{MRT}) * \mathcal{L}_{NLL}
\end{align}

We take the MRT training effect of the translation model optimized with BLEURT on En$\Rightarrow$De as an example to conduct experiments. 

\subsubsection{Results}
The results are shown in Figure \ref{fig:optimize} (b).
As can be seen, as the proportion of NLL loss increases, the decreasing trend of the remaining metrics gradually disappears.

The optimal result can be achieved when $\lambda_{MRT} = 0.6$ or $0.4$. At this point, unsupervised metrics remain stable, and supervised metrics show an increase.
This indicates that combining MRT and NLL loss can improve the training effect of the translation model.
For a fair comparison, we also check the results at the beginning of the optimization when using only MRT (before the training collapses). At this point, the improvement in BLEURT, COMET  and UniTE is more obvious, but accompanied by a decrease in BLEU and BERTScore. 
This suggests that the inclusion of NLL loss can make training more stable and more balanced across all metrics.

\subsection{Metrics Ensemble}
\subsubsection{Experiments}

Supervised metrics focus more on high-level semantic similarity and are considered to have a higher correlation with human evaluation \citep{kocmi2021ship}; while unsupervised metrics using word-level information are relatively stable and can ensure fine-grained text similarity \footnote{Note that although BARTScore is an unsupervised metric, it calculates the overall probability of sentence generation and still focuses more on sentence-level information.}.


We do an ensemble of different metrics in the hope that the integrated metrics can complement each other and integrate the advantages of different metrics. 
Then the ensemble metric is applied to MRT training on En$\Rightarrow$De. 

\subsubsection{Results}
\paragraph{Supervised and Unsupervised Metrics Ensemble.}
As can be seen in Figure \ref{fig:optimize} (a), optimizing BERTScore alone does not change the remaining metrics much, while only optimizing BLEURT reveals robustness problems. However, optimizing the ensemble of BERTScore and BLEURT works well: not only does it preserve the performance of the unsupervised metrics as much as possible, but it also leads to significant improvements in COMET and UniTE.

\paragraph{Supervised Metrics Ensemble.}
In addition, combining two sentence-level supervised metrics can also provide a boost, as the fifth column of Figure \ref{fig:optimize} (a) shows the effect of integrating BLEURT and COMET. 
Compared to optimizing only a single metric, 
we find that the ensemble metric can build on the strengths of both metrics.  While maintaining  the scores of  unsupervised metrics, it can further improve supervised metrics. COMET and UniTE all improve about 14.5\%, which is an increase of about 7 points. 
We suggest that this may be due to the fact that different metrics have different criteria for evaluating translation quality, and the robustness deficiency of one metric can be compensated by other metrics. 

\subsection{Method Validity Analysis}
\paragraph{Avoid High-Frequency Decoding Sentences.} 
We compare the entropy of decoded sentence frequencies on the En$\Rightarrow$De test set for the translation model trained with single or ensemble metrics.
As shown in Table \ref{tab:entropy}, the entropy is lower for the model trained with only BLEURT  because it decodes a large number of identical sentences.
While the frequency entropy for models trained with ensemble metrics is similar to that of the gold translations, indicating that the phenomenon of high-frequency decoded sentences disappears.

\paragraph{Comparison to MBR Decoding.} Minimum Bayes Risk (MBR) decoding can also get translations with metric style \citep{freitag2022high, muller2021understanding}.
Both MRT and MBR add some computational cost because they need to sample candidate translation sentences.
However, MRT is a training process that can quickly generate translations at test time once the model has finished training.
MBR, on the other hand, is a decoding process, which requires more time for each decoding. Therefore, from an application point of view, MRT is more efficient.


\begin{table}[]
\center
\small
\begin{tabular}{@{}cc@{}}
\toprule[1.2pt]
\textbf{System} & \textbf{Entropy} \\ \midrule
Ref & 11.55 \\
Hyp (Only BLEURT) & 6.58 \\
Hyp (BLEURT + BERTScore) & 11.55 \\
Hyp (BLEURT + COMET) & 11.55 \\ \bottomrule[1.2pt]
\end{tabular}
\caption{The entropy of decoded sentence frequencies on the En$\Rightarrow$De test set. Ref is the gold translation. Low entropy means that the translation model is damaged and decodes many identical sentences.}
\label{tab:entropy}
\end{table}

\section{Related Work}
\paragraph{Automatic Metrics}
Traditional metrics for machine translation evaluation including BLEU \citep{papineni2002bleu}, METEOR \citep{lavie2009meteor}, and chrF \citep{popovic2015chrf} are based on lexical overlap. 
Embedding-based metrics measure the semantic equivalence between the reference and translation hypothesis by contextual representation, such as BERTScore \citep{zhang2019bertscore}, MoverScore \citep{zhao2019moverscore}.
Generation-based metrics formulate the evaluation of  text as a generation task, such as BARTScore \citep{yuan2021bartscore} and PRISM \citep{thompson2020automatic}. The basic idea is that high quality text can be generated with high probability.
Learned metrics, such as BLEURT \citep{sellam2020bleurt}, COMET \citep{rei2020comet}, and the recently proposed UniTE \citep{wan2022unite} aim to train neural networks to directly predict human judgements. These supervised metrics correlate well with human evaluations, but lack interpretability and robustness studies, which is explored by this paper.

\paragraph{Minimum Risk Training}
\citet{shen2015minimum} proposes the MRT method and confirms its superiority with experiments.
\citet{edunov2017classical} compares various objective functions and further verifies that MRT training can enhance translation quality.
\citet{wang2020exposure} uses MRT to avoid exposure bias, thus improving translation quality in out-of-domain settings.
The above MRT work uses BLEU to guide the training of translation models, but BLEU is not the optimal metric. Our work uses various cutting-edge metrics to further improve translation quality.
\citet{wieting2019beyond} proposes a new metric, claiming its superiority over BLEU and suitability for MRT training. Our work, on the other hand, focuses on the analysis of metrics, with MRT serving as a tool to evaluate the robustness of various metrics systematically.

\paragraph{Metric Defects Analysis}
There are also some papers that start to explore the shortcomings of metrics. 
\citet{sai2021perturbation} provides perturbation templates to measure the performance of metrics on the constructed challenge set, while our work is to guide the metrics to generate adversarial samples (universal translations) by themselves.
\citet{amrhein2022identifying} does a case study on COMET through MBR decoding, showing that COMET is insensitive to numbers and named entities. 
Different from a pure case study, our work shows the tendency of metrics  through MRT, and can draw more typical conclusions.
\citet{sun2022bertscore} shows that PLM-based metrics, such as BERTScore, lack fairness and exhibit higher social bias than traditional metrics.
Our work analyzes metrics from a robustness perspective and  complements  this work.

\section{Conclusion}



In this paper, we present the first systematic analysis  of automatic metrics from the perspective of guidance for training machine translation systems. 
We find that MRT reveals the robustness deficiencies of some metrics, such as universal adversarial translations of BLEURT and BARTScore, and we further analyze the underlying causes. 
In addition, we explore methods to improve metric robustness, thus helping to further enhance the performance of translation systems.

\section*{Limitations}
First, we find robustness deficiencies in metrics by comparing the evaluation differences among metrics.
This applies to the case when there are metrics that do not have the same robustness flaws. If there are more latent common defects in the metrics, they cannot be identified by MRT.
We leave this topic for future research.

Second, we use beam search to generate candidates during MRT training, but beam search is also known to have deficiencies. For example, beam search suffers from heuristic search biases and shifts statistics away from those of the data \citep{eikema2020map}.
Different decoding methods may have an impact on the experiment results.



\section*{Acknowledgements}
We would like to thank the anonymous reviewers for their insightful comments. Shujian Huang is the corresponding author. This work is supported by National Science Foundation of China (No. 62176120), the Liaoning Provincial Research Foundation for Basic Research (No. 2022-KF-26-02).

\bibliographystyle{acl_natbib}
\bibliography{acl2023}

\clearpage
\appendix

\section{Ethics Statement}
This paper finds universal adversarial translations that can be used to attack metrics and lead to security risks. However, this paper also proposes methods to improve metric robustness to avoid this situation. 

\section{Training Duration for MLE and MRT}
\label{sec:train_duration}

We list the training duration for MLE and MRT in Table \ref{tab:train_epoch_num} and Table \ref{tab:train_step_num}, respectively. Table \ref{tab:train_epoch_num} shows the number of training epochs, while Table \ref{tab:train_step_num} shows the number of training steps. It can be seen that the training duration for MRT is much shorter than that for MLE.
The improvement of translation performance of the model mainly lies in the MLE stage, while MRT fine-tuning makes the model inclined towards specific metrics.

\section{MRT Training Process Figures}
\label{MRT Training Process Figures}
\begin{figure*}
  \centering
  \includegraphics[width=\textwidth]{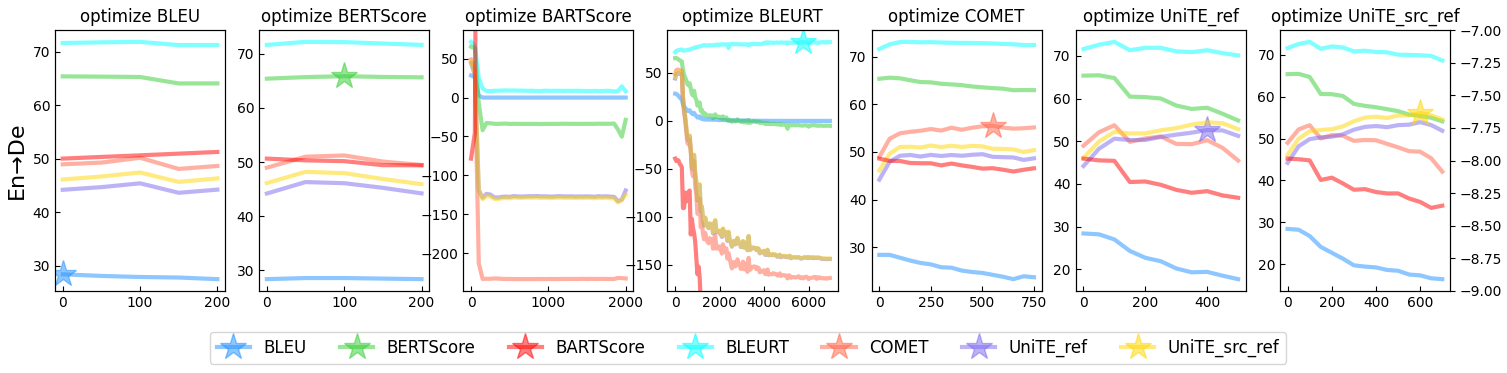}
  \caption{The training process of MRT optimized by each metric on En$\Rightarrow$De.}
  \label{fig:mrt_en2de}
\end{figure*}

\begin{figure*}
  \centering
  \includegraphics[width=\textwidth]{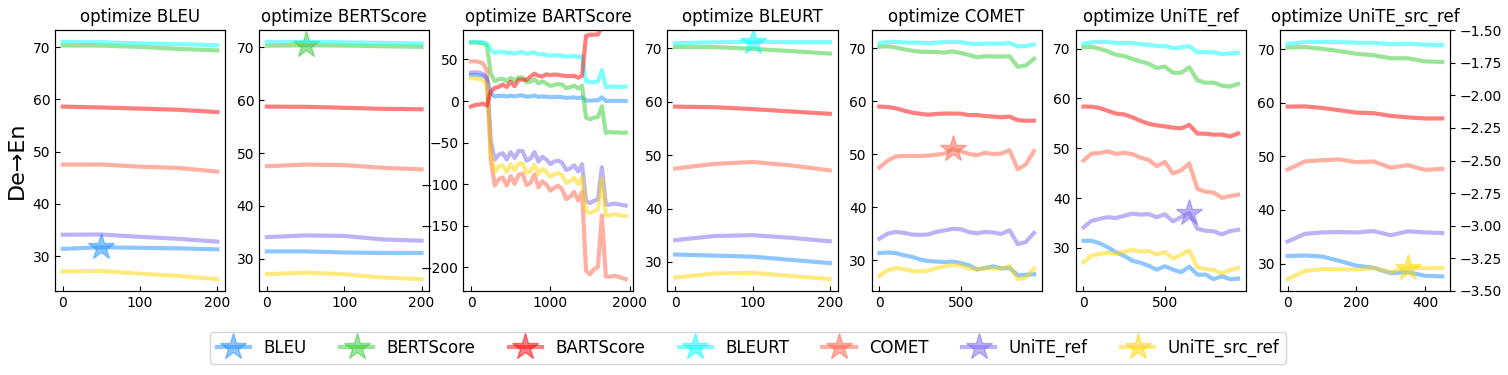}
  \caption{The training process of MRT optimized by each metric on De$\Rightarrow$En.}
  \label{fig:mrt_de2en}
\end{figure*}

\begin{figure*}
  \centering
  \includegraphics[width=\textwidth]{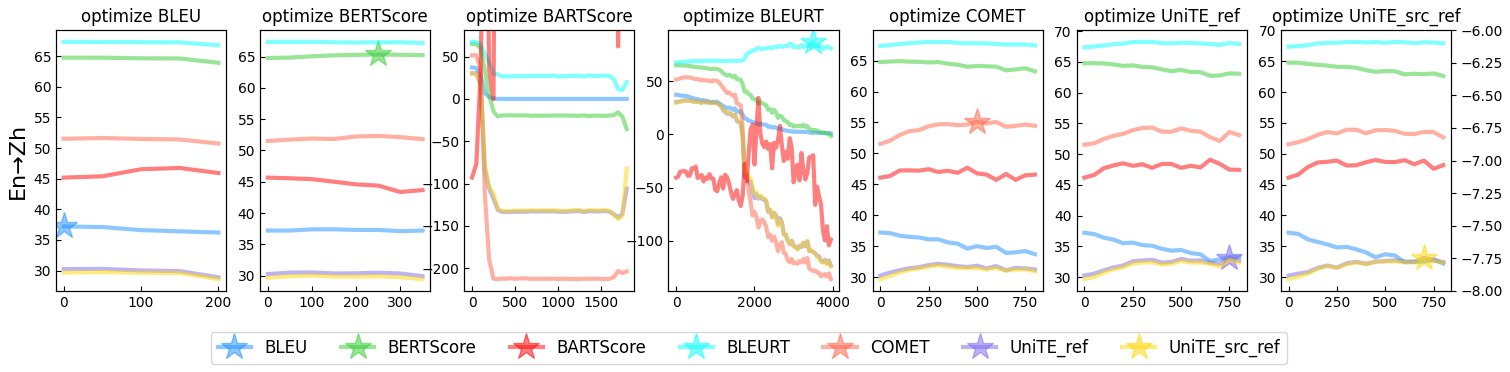}
  \caption{The training process of MRT optimized by each metric on En$\Rightarrow$Zh.}
  \label{fig:mrt_en2zh}
\end{figure*}

\begin{figure*}
  \centering
  \includegraphics[width=\textwidth]{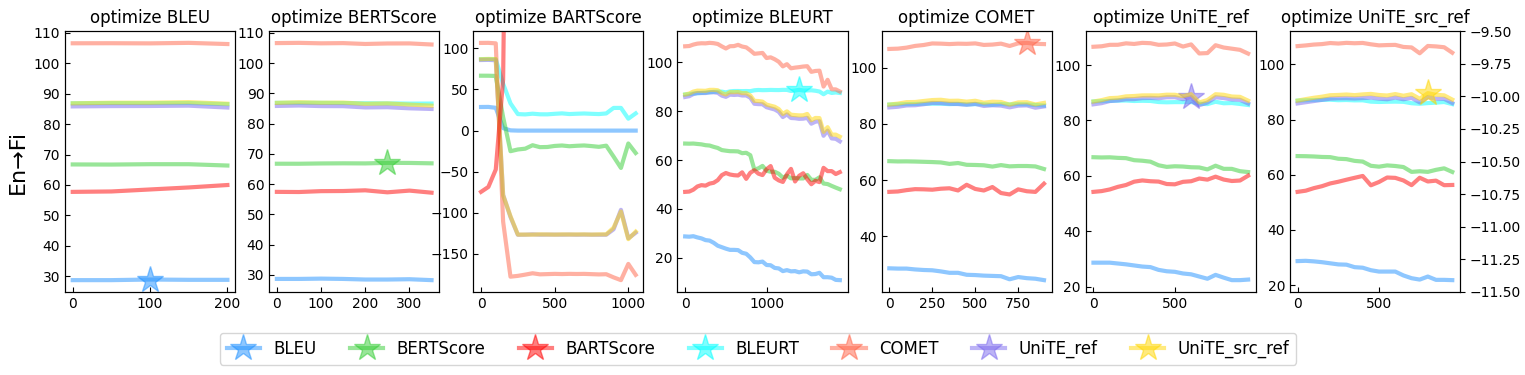}
  \caption{The training process of MRT optimized by each metric on En$\Rightarrow$Fi.}
  \label{fig:mrt_en2fi}
\end{figure*}

\begin{figure*}
  \centering
  \includegraphics[width=\textwidth]{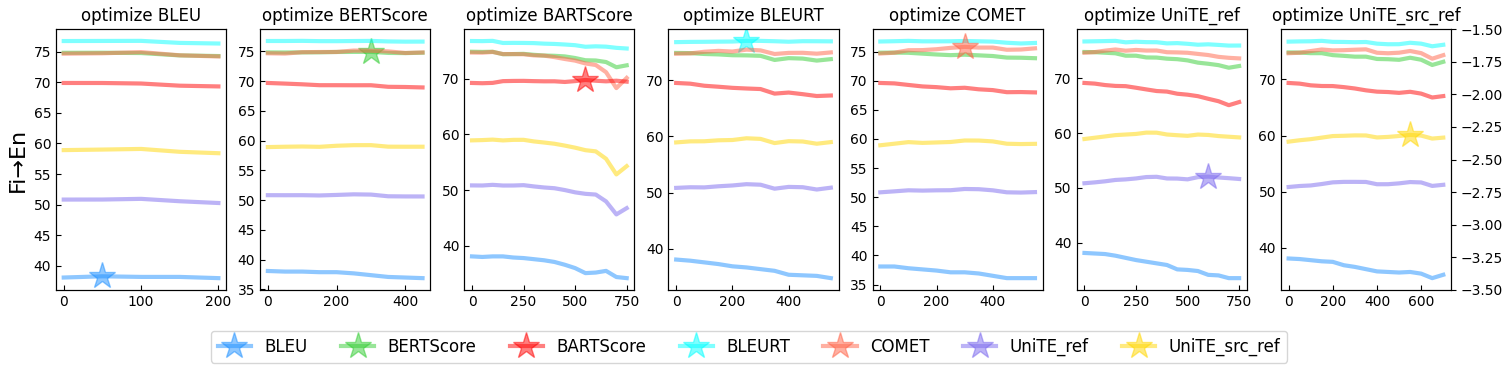}
  \caption{The training process of MRT optimized by each metric on Fi$\Rightarrow$En.}
  \label{fig:mrt_fi2en}
\end{figure*}

From Figure \ref{fig:mrt_en2de} to Figure \ref{fig:mrt_fi2en}, we can see how all metrics change when the translation model is optimized with each metric on different language pairs. From the trends of different metrics, we can observe the differences between the metrics and the impact of the metrics used for optimization on the translation model.

In each figure, the horizontal axis represents the training steps, and the vertical axis is the score of each metric  (except for BARTScore on the right axis, which is a negative number because it calculates the logarithmic probability of translations); metrics other than BARTScore and BLEU are mostly distributed between 0 and 1, and we multiply them uniformly by 100 for ease of observation. The asterisk represents the highest value achieved by the optimized metric.

\section{MRT Training Process Statistics}
\label{MRT Training Process Statistics}

Table \ref{tab:mrt_stat_de2en} to Table \ref{tab:mrt_stat_fi2en} display the change range in all metrics when optimizing the translation model with a specific metric to the highest point across different language pairs. The results correspond to figures in Appendix \ref{MRT Training Process Figures}.
$0.00\%$ means that the optimized metric does not continue to improve, and the highest value remains the same as the result of MLE training; a negative number means that the metric score goes from positive to negative, which means it decreases a lot.

\section{High Frequency Samples}
\label{high_freq}
Table \ref{tab:hotel} displays some hotel review examples in the WMT14 En$\Leftrightarrow$De dataset, and the semantics are very similar to universal translations of BLEURT on En$\Rightarrow$De. For ease of understanding, English is shown here.

\begin{table}[]
\begin{tabular}{p{0.9\columnwidth}}
\toprule[1.2pt]
\textbf{Examples of Hotel Review Sentences from WMT14 En$\Leftrightarrow$De} \\ \midrule
The location of the hotel was excellent. The room was clean and comfortable. \\ \midrule
The room was clean and comfortable, the hotel was situated close to the center but in the tourist center. The food was excellent and the service second to none. \\ \midrule
The location of the hotel is great, the atmosphere is quite pleasant, the staff is efficient and friendly, the room was clean and comfortable, the price was fair. In short words, everything was perfect. \\ \midrule
The room was clean and comfortable. \\ \midrule
the location of the hotel is ideal for sightseeing,the room was clean and comfortable, the staff were helpful. \\ \midrule
The room was clean and comfortable. Staff friendly. \\ \midrule
the employees were very helpful at all times the room was clean and comfortable and the restaurant was very nice. \\ \midrule
The room was clean and comfortable and the staff friendly and courteous. \\ \midrule
This is a great hotel .The room was clean and comfortable .With small budget but we have a comfortable stay .Good value, we will reccommend this hotel for anyone looking for a hotel in Hanoi . \\
\bottomrule[1.2pt]
\end{tabular}
\caption{Examples of hotel review sentences from WMT14 En$\Leftrightarrow$De.}
\label{tab:hotel}
\end{table}

\begin{table*}[]
\centering
\begin{tabular}{@{}cccccccc@{}}
\toprule[1.2pt]
\multicolumn{2}{c}{} & \textbf{En$\Rightarrow$De} & \textbf{De$\Rightarrow$En} & \textbf{En$\Rightarrow$Zh} & \textbf{Zh$\Rightarrow$En} & \textbf{En$\Rightarrow$Fi} & \textbf{Fi$\Rightarrow$En} \\ \midrule
\multicolumn{2}{c}{MLE} & 33 & 28 & 32 & 40 & 55 & 36 \\ \midrule
\multirow{7}{*}{MRT} & BLEU & 1 & 1 & 1 & 1 & 1 & 1 \\
 & BERTScore & 1 & 1 & 1 & 1 & 1 & 1 \\
 & BARTScore & 1 & 1 & 1 & 1 & 1 & 1 \\
 & BLEURT & 4 & 1 & 1 & 1 & 1 & 1 \\
 & COMET & 1 & 1 & 1 & 1 & 1 & 1 \\
 & UniTE\_ref & 1 & 1 & 1 & 1 & 1 & 1 \\
 & UniTE\_src\_ref & 1 & 1 & 1 & 1 & 1 & 1 \\ \bottomrule[1.2pt]
\end{tabular}
\caption{Comparison of the number of epochs trained by MLE and MRT. The number of epochs for MLE is the epoch number trained until early stop, while the number of epochs displayed in MRT is the epoch number when the model is optimized to the highest metric score.}
\label{tab:train_epoch_num}
\end{table*}
\begin{table*}[]
\centering
\begin{tabular}{@{}cccccccc@{}}
\toprule[1.2pt]
\multicolumn{2}{c}{} & \textbf{En$\Rightarrow$De} & \textbf{De$\Rightarrow$En} & \textbf{En$\Rightarrow$Zh} & \textbf{Zh$\Rightarrow$En} & \textbf{En$\Rightarrow$Fi} & \textbf{Fi$\Rightarrow$En} \\ \midrule
\multicolumn{2}{c}{Steps in one epoch} & 2403 & 2127 & 7927 & 7929 & 10906 & 10910 \\ \midrule
\multicolumn{2}{c}{MLE} & 163000 & 61000 & 51000 & 64000 & 126000 & 40000 \\ \midrule
\multirow{7}{*}{MRT} & BLEU & 0 & 50 & 0 & 200 & 100 & 50 \\
 & BERTScore & 100 & 50 & 250 & 200 & 250 & 300 \\
 & BARTScore & 1950 & 1900 & 1800 & 1450 & 1050 & 550 \\
 & BLEURT & 5750 & 100 & 3500 & 550 & 1400 & 250 \\
 & COMET & 550 & 450 & 500 & 550 & 800 & 300 \\
 & UniTE\_ref & 400 & 650 & 750 & 750 & 600 & 600 \\
 & UniTE\_src\_ref & 600 & 350 & 700 & 500 & 800 & 550 \\ \bottomrule[1.2pt]
\end{tabular}
\caption{Comparison of training steps between MLE and MRT. The number of steps for MLE is the number of steps trained until early stop, while the number of steps displayed in MRT is the number of steps when the model is optimized to the highest metric score.}
\label{tab:train_step_num}
\end{table*}

\begin{table*}[]
\small
\centering
\begin{tabular}{cccccccc}
\toprule[1.2pt]
\multirow{2}{*}{\textbf{Optimized Metric}} & \multicolumn{7}{c}{\textbf{Change Range of Metrics During MRT on De$\Rightarrow$En}} \\
 & BLEU & BERTScore & BARTScore & BLEURT & COMET & UniTE\_ref & UniTE\_src\_ref \\ \hline
BLEU & 0.96\% & -0.03\% & -0.32\% & -0.06\% & 0.04\% & 0.14\% & 0.41\% \\
BERTScore & 0.00\% & 0.12\% & -0.07\% & 0.05\% & 0.65\% & 0.93\% & 1.04\% \\
BARTScore & -99.68\% & -154.39\% & 39.02\% & -76.08\% & -547.63\% & -466.31\% & -608.53\% \\
BLEURT & -1.27\% & -0.50\% & -0.91\% & 0.36\% & 2.58\% & 2.72\% & 3.24\% \\
COMET & -5.10\% & -1.23\% & -2.57\% & 0.28\% & 7.41\% & 5.45\% & 7.84\% \\
UniTE\_ref & -16.56\% & -5.72\% & -6.71\% & -0.64\% & -1.13\% & 8.49\% & 8.76\% \\
UniTE\_src\_ref & -9.55\% & -2.85\% & -3.96\% & -0.01\% & 1.73\% & 5.57\% & 7.99\% \\
\bottomrule[1.2pt]
\end{tabular}
\caption{The change range of all metrics when one metric is optimized to the highest value during MRT on De$\Rightarrow$En.}
\label{tab:mrt_stat_de2en}
\end{table*}
\begin{table*}[]
\small
\centering
\begin{tabular}{cccccccc}
\toprule[1.2pt]
\multirow{2}{*}{\textbf{Optimized Metric}} & \multicolumn{7}{c}{\textbf{Change Range of Metrics During MRT on En$\Rightarrow$Zh}} \\
 & BLEU & BERTScore & BARTScore & BLEURT & COMET & UniTE\_ref & UniTE\_src\_ref \\ \hline
BLEU & 0.00\% & 0.00\% & 0.00\% & 0.00\% & 0.00\% & 0.00\% & 0.00\% \\
BERTScore & 0.27\% & 0.88\% & -0.86\% & -0.10\% & 1.55\% & 0.73\% & 0.76\% \\
BARTScore & -100.00\% & -155.17\% & 88.89\% & -70.57\% & -495.64\% & -450.90\% & -377.30\% \\
BLEURT & -96.77\% & -95.08\% & 2.40\% & 28.97\% & -349.43\% & -465.65\% & -472.03\% \\
COMET & -5.91\% & -1.00\% & 0.46\% & 0.67\% & 6.96\% & 5.27\% & 6.05\% \\
UniTE\_ref & -10.48\% & -2.54\% & 0.87\% & 1.03\% & 4.01\% & 9.58\% & 10.66\% \\
UniTE\_src\_ref & -12.63\% & -2.90\% & 1.83\% & 1.07\% & 3.90\% & 9.51\% & 11.34\% \\
\bottomrule[1.2pt]
\end{tabular}
\caption{The change range of all metrics when one metric is optimized to the highest value during MRT on En$\Rightarrow$Zh.}
\label{tab:mrt_stat_en2zh}
\end{table*}
\begin{table*}[]
\small
\centering
\begin{tabular}{cccccccc}
\toprule[1.2pt]
\multirow{2}{*}{\textbf{Optimized Metric}} & \multicolumn{7}{c}{\textbf{Change Range of Metrics During MRT on Zh$\Rightarrow$En}} \\
 & BLEU & BERTScore & BARTScore & BLEURT & COMET & UniTE\_ref & UniTE\_src\_ref \\ \hline
BLEU & 0.44\% & 0.26\% & 0.10\% & 0.29\% & 2.93\% & 4.32\% & 4.77\% \\
BERTScore & 0.44\% & 0.80\% & -0.10\% & 0.66\% & 6.05\% & 8.16\% & 10.24\% \\
BARTScore & -98.02\% & -126.46\% & 37.25\% & -44.95\% & -620.81\% & -825.23\% & -871.23\% \\
BLEURT & -7.05\% & -0.80\% & -1.86\% & 1.65\% & 11.99\% & 18.67\% & 20.28\% \\
COMET & -7.05\% & -0.98\% & -3.12\% & 0.77\% & 16.47\% & 16.50\% & 17.67\% \\
UniTE\_ref & -12.11\% & -2.45\% & -3.09\% & 0.61\% & 10.25\% & 26.77\% & 24.30\% \\
UniTE\_src\_ref & -8.81\% & -3.01\% & -5.14\% & -0.16\% & 11.75\% & 21.83\% & 27.31\% \\
\bottomrule[1.2pt]
\end{tabular}
\caption{The change range of all metrics when one metric is optimized to the highest value during MRT on Zh$\Rightarrow$En.}
\label{tab:mrt_stat_zh2en}
\end{table*}
\begin{table*}[]
\small
\centering
\begin{tabular}{cccccccc}
\toprule[1.2pt]
\multirow{2}{*}{\textbf{Optimized Metric}} & \multicolumn{7}{c}{\textbf{Change Range of Metrics During MRT on En$\Rightarrow$Fi}} \\
 & BLEU & BERTScore & BARTScore & BLEURT & COMET & UniTE\_ref & UniTE\_src\_ref \\ \hline
BLEU & 0.70\% & 0.12\% & 0.17\% & 0.05\% & -0.03\% & 0.17\% & 0.23\% \\
BERTScore & -0.70\% & 0.42\% & -0.03\% & -0.21\% & -0.12\% & -0.48\% & -0.21\% \\
BARTScore & -100.00\% & -140.82\% & 83.08\% & -75.93\% & -264.54\% & -244.66\% & -241.44\% \\
BLEURT & -51.22\% & -21.42\% & 1.14\% & 2.19\% & -8.02\% & -10.43\% & -9.85\% \\
COMET & -12.20\% & -2.58\% & 0.04\% & 0.07\% & 2.03\% & 0.86\% & 0.96\% \\
UniTE\_ref & -14.63\% & -5.43\% & 0.76\% & -0.21\% & 0.80\% & 3.00\% & 3.02\% \\
UniTE\_src\_ref & -19.51\% & -8.57\% & 0.74\% & -1.05\% & 0.05\% & 2.69\% & 3.11\% \\
\bottomrule[1.2pt]
\end{tabular}
\caption{The change range of all metrics when one metric is optimized to the highest value during MRT on En$\Rightarrow$Fi.}
\label{tab:mrt_stat_en2fi}
\end{table*}
\begin{table*}[]
\small
\centering
\begin{tabular}{cccccccc}
\toprule[1.2pt]
\multirow{2}{*}{\textbf{Optimized Metric}} & \multicolumn{7}{c}{\textbf{Change Range of Metrics During MRT on Fi$\Rightarrow$En}} \\
 & BLEU & BERTScore & BARTScore & BLEURT & COMET & UniTE\_ref & UniTE\_src\_ref \\ \hline
BLEU & 0.52\% & 0.01\% & -0.01\% & 0.02\% & 0.09\% & 0.01\% & 0.15\% \\
BERTScore & -1.84\% & 0.13\% & -0.91\% & 0.02\% & 0.58\% & 0.20\% & 0.55\% \\
BARTScore & -7.87\% & -2.04\% & 1.20\% & -1.34\% & -2.63\% & -2.97\% & -2.99\% \\
BLEURT & -3.67\% & -0.58\% & -2.22\% & 0.29\% & 0.98\% & 1.30\% & 1.25\% \\
COMET & -2.62\% & -0.46\% & -1.92\% & 0.11\% & 1.53\% & 1.12\% & 1.39\% \\
UniTE\_ref & -10.50\% & -2.88\% & -6.33\% & -0.68\% & -0.55\% & 2.39\% & 1.25\% \\
UniTE\_src\_ref & -6.30\% & -1.26\% & -3.53\% & -0.26\% & 0.53\% & 1.73\% & 2.05\% \\
\bottomrule[1.2pt]
\end{tabular}
\caption{The change range of all metrics when one metric is optimized to the highest value during MRT on Fi$\Rightarrow$En.}
\label{tab:mrt_stat_fi2en}
\end{table*}

\end{CJK}
\end{document}